\documentclass{llncs}

\usepackage{url}

\usepackage{amsmath}

\usepackage{graphicx}

\usepackage{mdframed}

\usepackage{float}

\usepackage{etoolbox}
\newtoggle{submission}
\newtoggle{arxiv}
\toggletrue{arxiv}

\iftoggle{submission}{%
}{%
\usepackage{xcolor,pagecolor}

\iftoggle{arxiv}{%
}{%
\definecolor{background_color}{HTML}{1E1E1E}
\definecolor{text_color}{HTML}{D4D4D4}
\pagecolor{background_color}
\color{text_color}
}

\usepackage{hyperref}
\usepackage{fancyhdr}
\fancypagestyle{copyright}{\fancyhf{}\fancyfoot[L]
{\textcopyright{ }Springer Nature Switzerland AG 2020 \\
Z. Chen et al. (Eds.): ICBC 2020, LNCS 12404, pp. 1–12, 2020. \\
The final authenticated version is available online at \url{https://doi.org/10.1007/978-3-030-59638-5_10}.
}}
}

\begin{document}

\title{Analysis of Models for Decentralized and Collaborative AI on Blockchain}

\author{Justin D. Harris}
\institute{Microsoft, Montreal/Toronto, Canada \\
    \email{justin.harris@microsoft.com}
}

\maketitle

\begin{abstract}
	Machine learning has recently enabled large advances in artificial intelligence, but these results can be highly centralized.
	The large datasets required are generally proprietary; predictions are often sold on a per-query basis; and published models can quickly become out of date without effort to acquire more data and maintain them.
	Published proposals to provide models and data for free for certain tasks include Microsoft Research's Decentralized and Collaborative AI on Blockchain.
	The framework allows participants to collaboratively build a dataset and use smart contracts to share a continuously updated model on a public blockchain.
	The initial proposal gave an overview of the framework omitting many details of the models used and the incentive mechanisms in real world scenarios.
	For example, the Self-Assessment incentive mechanism proposed in their work could have problems such as participants losing deposits and the model becoming inaccurate over time if the proper parameters are not set when the framework is configured.
	In this work, we evaluate the use of several models and configurations in order to propose best practices when using the Self-Assessment incentive mechanism so that models can remain accurate and well-intended participants that submit correct data have the chance to profit.
	We have analyzed simulations for each of three models: Perceptron, Naïve Bayes, and a Nearest Centroid Classifier, with three different datasets: predicting a sport with user activity from Endomondo, sentiment analysis on movie reviews from IMDB, and determining if a news article is fake.
	We compare several factors for each dataset when models are hosted in smart contracts on a public blockchain: their accuracy over time, balances of a good and bad user, and transaction costs (or gas) for deploying, updating, collecting refunds, and collecting rewards.
	A free and open source implementation for the Ethereum blockchain and simulations written in Python is provided at \url{https://github.com/microsoft/0xDeCA10B}.
	This version has updated gas costs using newer optimizations written after the original publication.
\end{abstract}


\nottoggle{submission}{%
\thispagestyle{copyright}
}

\section{Introduction}
The advancement of popular blockchain based cryptocurrencies such as Bitcoin \cite{nakamoto2008bitcoin} and Ethereum \cite{buterin2015ethereum} have inspired research in decentralized applications that leverage these publicly available resources.
One application that can greatly benefit from decentralized public blockchains is the collaborative training of machine learning models to allow users to improve a model in novel ways \cite{harris2019decentralized}.
There exists several proposals to use blockchain frameworks to enable the sharing of machine learning models.
In DInEMMo, access to trained models is brokered through a marketplace allowing contributors to profit based on a model's usage, but it limits access to just those who can afford the price \cite{Marathe2018DInEMMoDI}.
DanKu proposes a framework for competitions by storing already trained models in smart contracts, which do not allow for continual updating \cite{danku_protocol}.
Proposals to change Proof-of-Work (PoW) to be more utilitarian by training machine learning models have also gained in popularity such as: A Proof of Useful Work for Artificial Intelligence on the Blockchain \cite{lihu2020proof}.
These approaches can incite more technical centralization such as harboring machine learning expertise, siloing proprietary data, and access to machine learning model predictions (e.g. charged on a per-query basis).
In the crowdsourcing space, a decentralized approach called CrowdBC has been proposed to use a blockchain to facilitate crowdsourcing \cite{8540048}.

To address centralization in machine learning, frameworks to share machine learning models on a public blockchain while keeping the models free to use for inference have been proposed.
One example is Decentralized and Collaborative AI on Blockchain from Microsoft Research \cite{harris2019decentralized}.
That work focuses on the description of several possible incentive mechanisms to encourage participants to add data to train a model.
This is the continuation of previous work in \cite{harris2019decentralized} by the author.

The system proposed in \cite{harris2019decentralized} is modular: different models or incentive mechanisms (IMs) can be used an seamlessly swapped: however, some IMs might work better for different models and vice-versa.
These models can be efficiently updated with one sample at time making them useful for deployment on Proof-of-Work (PoW) blockchains \cite{harris2019decentralized} such as the current public Ethereum \cite{buterin2015ethereum} blockchain.
The first is a Naive Bayes classifier for its applicability to many types of problems \cite{naiveBayes}.
Then, a Nearest Centroid Classifier \cite{Tibshirani6567ncc}.
Finally, a single layer Perceptron model \cite{rosenblatt1958perceptron}.

We evaluated the models on three datasets that were chosen as examples of problems that would benefit from collaborative scenarios where many contributors can improve a model in order to create a shared public resource.
The scenarios were: predicting a sport with user activity from Endomondo \cite{endomondo-10.1145/3308558.3313643}, sentiment analysis on movie reviews from IMDB\cite{maas-EtAl:2011:ACL-HLT2011}, and determining if a news article is fake \cite{FakeNewsKaggle}.
In all of these scenarios users benefit from having a direct impact on improving a model they frequently use and not relying on a centralized authority to host and control the model. 
Transaction costs (or gas) for each operation were also compared since these costs can be significant for the public Ethereum blockchain.

The Self-Assessment IM allows ongoing verification of data contributions without the need for a centralized party to evaluate data contributions.
Here are the highlights of the IM as explained in \cite{harris2019decentralized}:
\begin{itemize}
	\item \emph{Deploy}:
	One model, $h$, already trained with some data is deployed.
	\item \emph{Deposit}:
	Each data contribution with data $x$ and label $y$ also requires a deposit, $d$.
	Data and meta-data for each contribution is stored in a smart contract.
	\item \emph{Refund}:
	To claim a refund on their deposit, after a time $t$ has passed and if the current model, $h$, still agrees with the originally submitted classification, i.e. if $h(x) == y$, then the contributor can have their entire deposit $d$ returned.
	\begin{itemize}
		\item We now assume that $(x, y)$ is ``verified'' data.
		\item The successful return of the deposit should be recorded in a tally of points for the wallet address.
	\end{itemize}
	\item \emph{Take}:
	A contributor that has already had data verified in the \emph{Refund} stage can locate a data point $(x,y)$ for which $h(x) \neq y$ and request to take a portion of the deposit, $d$, originally given when $(x,y)$ was submitted.
\end{itemize}

If the sample submitted, $(x,y)$ is incorrect, then within time $t$, other contributors should submit $(x, y')$ where $y'$ is the correct or at least generally preferred label for $x$ and $y' \neq y$.
This is similar to how one generally expects bad edits to popular Wikipedia \cite{wikipedia} articles to be corrected in a timely manner.

As proposed, the Self-Assessment IM could result in problems such as participants losing deposits and the model becoming inaccurate if the proper parameters are not set when the framework is initially deployed.
In this work, we analyze the choice of several possible supervised models and configurations with the Self-Assessment IM in order to find best practices.

\section{Machine Learning Models} \label{sec:models}
In this section, we outline several models choices of machine learning model for use with Decentralized and Collaborative AI on Blockchain as proposed in \cite{harris2019decentralized}.
The model architecture chosen relates closely to the incentive mechanism chosen.
In this work, we will analyze models for the Self-Assessment incentive mechanism as it appeals to the decentralized nature of public blockchains in that a centralized organization should not need to maintain the IM, for example, by funding it \cite{harris2019decentralized}.

For our experiments, we mainly consider supervised classifiers because they can be used for many applications and can be easily evaluated using test sets.
In order to keep transaction costs low, we first propose to leverage the work in the Incremental Learning space \cite{Schlimmer:1986:CSI:2887770.2887853} by using models capable of efficiently updating with one sample.
Transaction costs, or ``gas'' as it is called in Ethereum \cite{buterin2015ethereum}, are important for most public blockchains as a way to pay for the computation cost for executing a smart contract.

\subsection{Naive Bayes}
The model first is a Naive Bayes classifier for its applicability to many types of problems \cite{naiveBayes}.
The Naive Bayes classifier assumes each feature in the model is independent, this is what helps makes computation fast when updating and predicting.
To update the model, we just need to update several counts such as the number of data points seen, the number of times each feature was seen, the number of times each feature was seen for each class, etc.
When predicting, all of these counts are used for the features presented in the sparse sample to compute the most likely class for the sample using Bayes' Rule \cite{naiveBayes}.

\subsection{Nearest Centroid}
A Nearest Centroid Classifier computes the average point (or centroid) of all points in a class and classifies new points by the label of the centroid that they are closest to \cite{Tibshirani6567ncc}.
They can also be easily adapted to support multiple classifications (which we do not do for this work).
For this model, we keep track of the centroid for each class and update it using the cumulative moving average method \cite{wikipedia:Moving_avg}.
Therefore we also need to record the number of samples that have been given for each class.
Updating the model with one sample needs to update the centroid for the given class but not for the other classes.
This model can be used with dense data representations.

\subsection{Perceptron}
A single layer perceptron model is useful linear model for binary classification \cite{rosenblatt1958perceptron}.
We evaluate this model because it can be used for sparse data like text as well as dense data.
The Perceptron's update algorithm only updates the weights if the model currently classifies the sample as incorrect.
This is good for our system since it should help avoid overfitting.
The model can be efficiently updated by just adding or subtracting, depending on the sample's label, the values for the features of the sample with the model's weights.

\section{Datasets}
Three datasets were chosen as examples of problems that would benefit from collaborative scenarios where many contributors can improve a model in order to create a shared public resource.
In each scenario, the users of an application that would use such a model benefit by having a direct impact on improving the model they frequently use and not relying on a centralized authority to host and control the model.

\subsection{Fake News Detection}
Given the text for a news article, the task is the determine if the story is reliable or not \cite{FakeNewsKaggle}.
We convert each text to a sparse representation using the term-frequency of the bigrams with only the top 1000 bigrams by frequency count in the training set considered.
While solving fake news detection is likely too difficult for simple models, a detector would greatly benefit from decentralization: freedom from being biased by a centralized authority.

\subsection{Activity Prediction}
The FitRec datasets contain information recorded from the use of participants' fitness trackers during certain activities \cite{endomondo-10.1145/3308558.3313643}.
In order to predict if someone was biking or running, we used the following features: heart rate, maximum speed, minimum speed, average speed, median speed, and gender.
We did some simple feature engineering with those features such as using average heart rate divided by minimum heart rate.
As usual, all of our code is public.

Fitness trackers and start-ups developing them have gained in popularity in recent years.
A user considering purchasing a new tracker might not trust that the manufacturer developing it will still be able to host a centralized model in few years.
The company could go bankrupt or just discontinue the service.
Using a decentralized model gives users confidence that the model they need will be available for a long time, even if the company is not.
This should even give them the assurance to buy the first version of a product and knowing that it should improve without them getting forced into buying a later version of the product.
Even if the model does get corrupted, applications can easily revert to an earlier version on the blockchain, still giving users the service they need \cite{harris2019decentralized}.

\subsection{IMDB Movie Review Sentiment Analysis}
The dataset of 25,000 IMDB movie reviews from is a dataset for sentiment analysis where the task is to predict if the English text for a movie review is positive or negative \cite{maas-EtAl:2011:ACL-HLT2011}.
We used word-based features limited to only the most 1000 common words in the dataset.
This particular sentiment analysis dataset was chosen for this work because of it's size and popularity.
Even though this dataset focuses on movie reviews, in general, a collaboratively built model for sentiment analysis can be applicable in many scenarios such as a system to monitor posts on social media.
Users could train a shared model when they flag posts or messages as abusive and this model can be used by several social media services to provide a more pleasant experience to their users.

\section{Experiments} \label{sec:experiments}
We conducted experiments for the three datasets with each of the three models.
Experiments ran in simulations to quickly determine the outcome of different configurations.
The code for our simulations is all public.
Each simulation starts with a model trained on 10\% of the training data.
The simulation then iterates over the rest of the samples in the training set submitting each sample once.

For simplicity, we assumed that each scenario just has two agents representing the main two types of user groups: ``good'' and ``bad''.
We refer to these as agents since they may not be real users but could be programs possibly even generating data to submit.
The ``good'' agent almost always submits correct data with the label as provided in the dataset, as a user would normally submit correct data in a real-world use case.
The ``bad'' agent represents those that wish to decrease the model's performance, so the ``bad'' agent always submits data with the opposite label that was provided in the dataset.
Since the ``bad'' agent is trying to corrupt the model, they are willing deposit more (when required) to update the model.
This allows them to update the model more quickly after the model has already been updated.
The ``good'' agent only updates the model if the deposit required to do so is low, otherwise they will wait until later.
They also check the model's recent accuracy on the test set before submitting data.
In the real world, it is important for people to monitor if the model's performance and determine if it is worth trying to improve it or if it is totally corrupt.
If the model's accuracy is around 85\% then it can be assumed to be okay and not overfitting so ideally, it should be safe to submit new data.
If incorrect data was always submitted, or submitted too often by ``bad'' agents, then of the model's accuracy should decrease and honest users would most likely lose their deposits because their data would not satisfy the refund criteria of the IM.
We use loose terms here like ``should'' and ``likely'' because it is difficult to be general in terms of all types of models.
For example, certainly a rule-based model could be used that memorizes training data.
As long as no duplicate data is submitted with different labels, a rule-based model would allow each participant to get their deposits back and the analysis would be trivial.
The characteristics of the agents are compared in Table \ref{table:agent_table}.

\begin{table}
	\caption{Characteristics of the agents' behaviors}\label{table:agent_table}
	\begin{center}	
		\begin{tabular}{lll}
			\hline\noalign{\smallskip}
			Characteristic & ``Good'' Agent & ``Bad'' Agent \\
			\noalign{\smallskip}
		\hline
		\noalign{\smallskip}
		Starting Balance & 10,000 & 10,000 \\
		Average Maximum Deposit & 50 & 100 \\
		Deposit Standard Deviation & 10 & 3 \\
		Average Time Between Updates & 10 minutes & 60 minutes \\  
		P(incorrect label) & 0.01\%  & 100\% \\
		P(submitting) & $(100 * \text{accuracy} + 15)$\%  & 100\% \\
		\hline
		\end{tabular}
	\end{center}
\end{table}

Each agent must wait 1 day before being claiming a refund for ``good'' data or reporting the data as ``bad''.
This was referred to as $t$ in our original paper.
When reporting data as ``bad'', an agent can an amount from the initial deposit portional to the percent of ``verified'' contributions they have.
This can be written as $r(c_r, d) = d \times \frac{n(c_r)}{\sum_{\text{all } c} n(c)}$ using the notation in our initial paper.
After 9 days, either agent can claim the entire remaining deposit for a specific data sample.
This was $t_a$ in our original paper.

For each dataset, we compared:
\begin{itemize}
	\item The change of each agent's \emph{balance} over time.
	While using the IM, an agent may lose deposits to the other agent, reclaim their deposit, or profit by taking deposits that were from the other agent.
	We monitor balances in order to determine if it can be beneficial for an agent to participate by submitting data, whether it be correct or incorrect.
	\item The change of the model's \emph{accuracy} with respect to a fixed test set over time.
	In a real-world scenario, it would be important for user's to monitor the accuracy as a proxy to measure if they should continue to submit data to the model.
	If the accuracy declines, then it could mean that ``bad'' agents have corrupted the model.
	\item The ``ideal'' baseline of the model's accuracy on the test set if the model were to be trained all of the simulation data.
	In the real-world, this would of course not be available because the data would not be known yet.
\end{itemize}

We also compared Ethereum gas costs (i.e. transaction costs) for the common actions that are done in the framework.
The Update gas cost shown for each model was when the model did not agree with the provided label classification and so needed its weights to be updated.
Otherwise, the Perceptron Update method would be only slightly more than prediction because a Perceptron model does not get updated if it currently predicts the same classification as the label it is given for a data sample.
The gas cost of predicting is not shown because it can be done ``off-chain'' (without creating a transaction) which incurs no gas cost since it does not involve writing data to the blockchain.
However, predicting is the most expensive operation inside of Refund and Report so the cost of doing prediction ``on-chain'' can be estimated using those operations.
Contracts were compiled with the ``solc-js'' compiler using Solidity 0.6.2.

\subsection{Fake News Detection}
\begin{figure}
	\centering
	\includegraphics[width=0.9\textwidth]{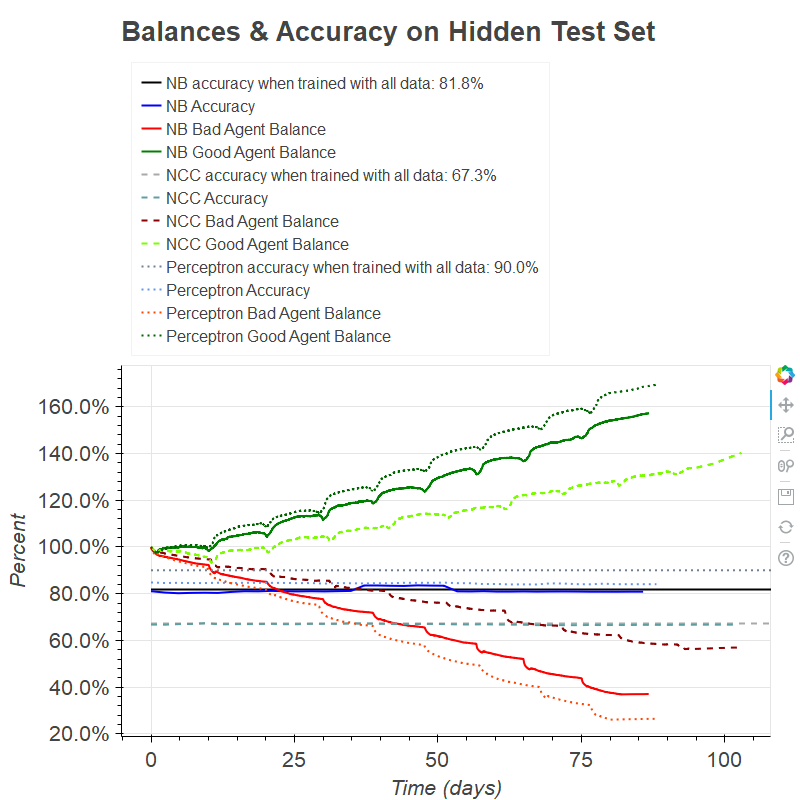}
	\caption{
		Plot of simulations with the Fake News dataset.
	}\label{figure:news_plot}
\end{figure}

With each model, the ``good'' agent was able to profit and the ``bad'' agent lost funds.
As can be seen in Figure \ref{figure:news_plot}, the difference in balances was most significant with the Perceptron model.
The Perceptron model has the highest accuracy yet the Naive Bayes was able to surpass its baseline accuracy.

The Perceptron model has the lowest gas cost as shown in Table \ref{table:news_gas_costs}.
The deployment cost for the Naive Bayes and Sparse Nearest Centroid models were much higher because almost all of the 1000 features effectively needs to be set twice (once for each class).
For the Sparse Nearest Centroid Classifier, prediction (which happens in Refund and Reward) did not need to go through each dimension because the distance to each centroid can be calculated by storing the magnitude of each centroid and then using the sparse input data to find the difference from the magnitude just for the few features in the sparse input.
Updating the Sparse Nearest Centroid Classifier does not need to update every feature because we store some extra information (mainly the new denominator) when we update each feature.
At prediction time, we compute use the correct denominator to use.

\begin{table}
	\caption{
		Ethereum gas costs for each model for the Fake News dataset.
		Data samples had 15 integer features representing the presence of the top bigrams from the training data.
		In brackets are approximate USD values from September 2020 with a modest gas price of 4gwei and ETH valued at 373 USD.
	}\label{table:news_gas_costs}
	\begin{center}	
		\begin{tabular}{llll}
			\hline\noalign{\smallskip}
			Action & Naive Bayes & Sparse Nearest Centroid & Sparse Perceptron \\
			\noalign{\smallskip}
		\hline
		\noalign{\smallskip}
		Deployment & 55,511,446 & 67,139,037 & \textbf{30,967,145} (46.20 USD) \\
		Update & 281,447 & 356,345 & \textbf{263,517} (0.39 USD) \\
		Refund & 172,216 & 176,797 & \textbf{138,028} (0.21 USD) \\
		Reward & 136,800 & 141,253 & \textbf{102,484} (0.15 USD) \\
		\hline
		\end{tabular}
	\end{center}
\end{table}

\subsection{Activity Prediction}
\begin{figure}
	\centering
	\includegraphics[width=0.9\textwidth]{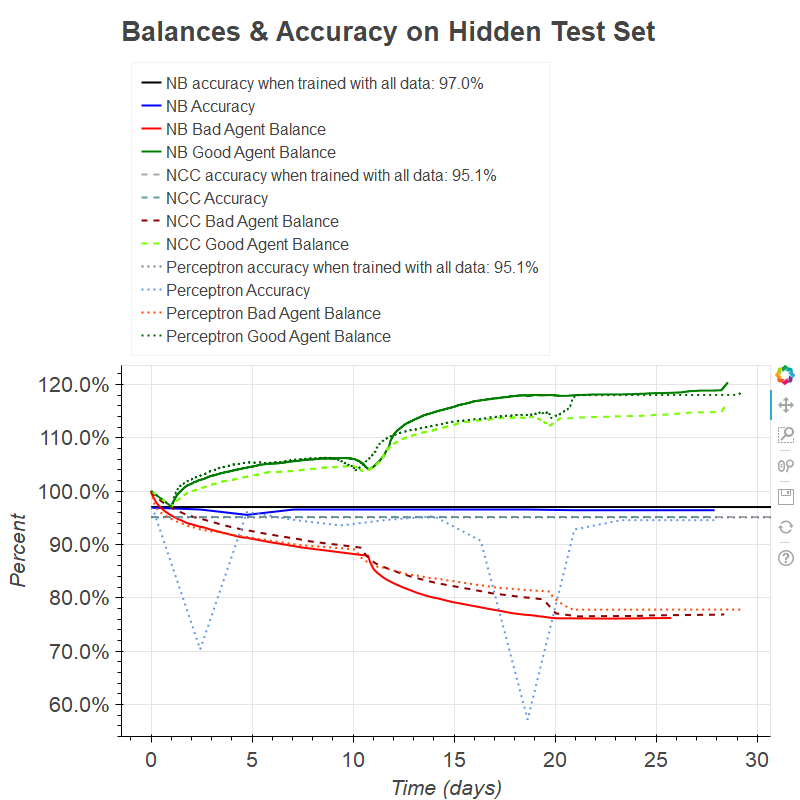}
	\caption{
		Plot of simulations with the Activity Prediction dataset.
	 }
	\label{figure:fitness_plot}
\end{figure}

As seen in Figure \ref{figure:fitness_plot}, with each model, all ``good'' agents can profit while the ``bad'' agent wastes lots of funds.
The Naive Bayes (NB) and Nearest Centroid Classifier (NCC) models performed very well on this type of data, hardly straying from the ideal baseline.
The linear Perceptron on the other hand was much more sensitive to data from the ``bad'' agent and it's accuracy dropped significantly several times but finally recovering.
This could be because the Percepton does not update if it already agrees with the classification provided.
So it might not have been able to gain as much reinforcement from correct data as the other classifiers.

The Perceptron model usually has the lowest gas cost as shown in Table \ref{table:fitness_gas_costs}.
The gas costs were fairly close for each action amongst the models, especially compared to the other datasets.
This is mostly because there are very few features (just 9) for this dataset.
Dense versions of the models are very expensive when you have many features.

\begin{table}
    \caption{
        Ethereum gas costs for each model for the Activity Prediction dataset.
        Data samples had 9 integer features.
        In brackets are approximate USD values from September 2020 with a modest gas price of 4gwei and ETH valued at 373 USD.
    }\label{table:fitness_gas_costs}
	\begin{center}	
		\begin{tabular}{llll}
			\hline\noalign{\smallskip}
			Action & Naive Bayes & Dense Nearest Centroid & Dense Perceptron \\
			\noalign{\smallskip}
		\hline
		\noalign{\smallskip}
		Deployment & 10,113,606 & 9,734,985 & \textbf{8,977,816} (13.39 USD) \\
		Update & \textbf{222,523} (0.33 USD) & 243,164 & 227,047 \\
		Refund & 151,070 & 146,790 & \textbf{133,745} (0.20 USD) \\
		Reward & 115,525 & 111,245 & \textbf{98,238} (0.15 USD) \\
		\hline
		\end{tabular}
	\end{center}
\end{table}

\subsection{IMDB Movie Review Sentiment Analysis}
\begin{figure}
	\centering
	\includegraphics[width=0.9\textwidth]{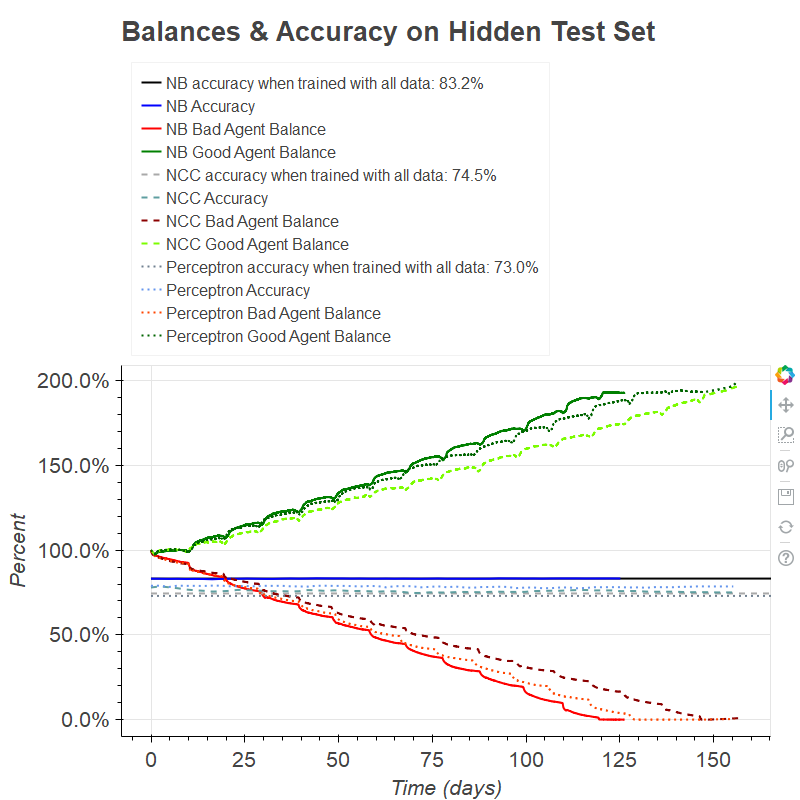}
	\caption{Plot of simulations with the IMDB Movie Review Sentiment Analysis dataset.}
	\label{figure:imdb_plot}
\end{figure}

Figure \ref{figure:imdb_plot} shows all ``good'' agents can profit while the ``bad'' agent loses most or all of the initial balance.
All models maintained their accuracy with this type of data with the Naive Bayes model performing the best.
This is likely because there are so many features.

By only a small amount, the Naive Bayes model beats the Perceptron model for the Update method with the lowest gas cost.
The gas costs for all actions are shown in Table \ref{table:imdb_gas_costs}.
As with the Fake News dataset, the Update cost for the Sparse Nearest Centroid Classifier was low because we can skip many dimensions.
The Sparse Nearest Centroid Classifier and Naive Bayes models had a much higher deployment costs since the amount of data was effectively double since each feature needs to be set for each of the two classes.

\begin{table}
	\caption{
		Ethereum gas costs for each model for the IMDB Movie Review Sentiment Analysis dataset.
		Data samples had 20 integer features representing a movie review with the presence of 20 words that were in the 1000 most common words in the training data.
		In brackets are approximate USD values from September 2020 with a modest gas price of 4gwei and ETH valued at 373 USD.
	}\label{table:imdb_gas_costs}
	\begin{center}	
		\begin{tabular}{llll}
			\hline\noalign{\smallskip}
			Action & Naive Bayes & Sparse Nearest Centroid & Sparse Perceptron \\
			\noalign{\smallskip}
		\hline
		\noalign{\smallskip}
		Deployment & 55,423,682 & 67,136,669 & \textbf{30,875,193} (46.07 USD) \\
		Update & \textbf{332,636} (0.50 USD) & 422,476 & 332,927 \\
		Refund & 189,954 & 196,375 & \textbf{145,601} (0.22 USD) \\
		Reward & 154,538 & 160,831 & \textbf{110,157} (0.16 USD) \\
		\hline
		\end{tabular}
	\end{center}
\end{table}

\section{Conclusion}
With all experiments, the Perceptron model was consistently the cheapest to use.
This was mostly because the size of the model was much less than the other two models which need to store information for each class, effectively twice the amount of information that the Perceptron needs to store.
While each model was expensive to deploy, this is a one time cost to incur.
This cost is far less than the comparable cost to host a web service with the model for several months.

Most models were able to maintain their accuracy except for the volatile Perceptron for the Activity Prediction dataset.
Even if the model gets corrupted with incorrect data, it can be forked from an earlier time when its accuracy on a hidden test set was higher.
It can also be retrained with data identified as ``good'' while it was deployed.
It is important for users to be aware of the accuracy on the model on some hidden test set.
Users can maintain their own hidden test sets or possibly use a service supplied by an organization which would publish a rating on a model based on the test sets they have.

The balance plots looked mostly similar across the experiments because the ``good'' agent was already careful and how we set a constant wait time of 9 days for either agent to claim the remaining deposit for a data contribution.
The ``good'' agent honestly submitted correct data and only did so when they thought the model was reliable, this helped ensure that they can recover their deposits and earn for reporting many contributions from the ``bad agent''.
When the ``bad'' agent is able to corrupt, it can successfully report a portion of the contributions from the ``good'' agent as bad because the model would not agree with those contributions.
The ``bad'' agent cannot claim a majority of these deposits when reporting the contribution since they do not have as many ``verified'' contributions as the ``good'' agent.
This leaves a left over amount for which either agent must wait for 9 days before taking the entire remaining deposit, hence the periodic looking patterns in the balance plots every 9 days.
The pattern continues throughout the simulation because there is always data for which the deposit that cannot be claimed by either agent after just the initial refund wait time of 1 day.

Future work in analyzing more scenarios is encouraged and easy to implement with our open source tools at \url{https://github.com/microsoft/0xDeCA10B/tree/master/simulation}. For example, changing the initial balances of each agent to determine how much a ``good'' agent need to spend to stop a much more resourceful ``bad'' agent willing to corrupt a model.

\bibliographystyle{splncs}
\bibliography{abbreviations,citations}

\end{document}